\documentclass[10pt,twocolumn,letterpaper]{article}
\usepackage{cvpr}
\usepackage{times}
\usepackage{epsfig}
\usepackage{graphicx}
\usepackage{amsmath}
\usepackage{amssymb}
\usepackage{amsmath}
\usepackage{amssymb}
\usepackage{relsize}
\usepackage{tabularx}
\usepackage{booktabs}
\newcommand{\ra}[1]{\renewcommand{\arraystretch}{#1}}
\usepackage{multirow}
\usepackage{siunitx} %
\usepackage{dcolumn}
\newcolumntype{d}[1]{D{.}{.}{#1}}
\usepackage{relsize}
\usepackage{xcolor}
\newcommand{\mc}[1]{\multicolumn{1}{c}{#1}}

\newcommand*\mystrut[1]{\vrule width0pt height0pt depth#1\relax}  %
\newcommand{\norm}[1]{\Big\lVert#1\Big\rVert}  %
\newcommand{\realset}{\ensuremath{ \mathbb{R}}}   %
\newcommand{\dataset}{\ensuremath{ \mathbb{D}}}   %
\usepackage[scr]{rsfso}  %
\newcommand{\Laplace}{\mathscr{L}}  %
\newcommand{\loss}{\Laplace}
\newcommand{\hd}{\ensuremath{ d_{\mathcal{H}}}}  %
\newcommand{\xreal}{\ensuremath{ {\bf x}}}   %
\newcommand{\xrealvec}{\ensuremath{ \vec{\xreal} }}   %
\newcommand{\xrealbatch}{\ensuremath{ {\bf X}}}   %
\newcommand{\X}{\ensuremath{ {\bf X}}}   %
\newcommand{\Y}{\ensuremath{ {\bf Y}}}   %
\newcommand{\xfake}{\ensuremath{ {\bf x'}}}  %
\newcommand{\xfakevec}{\ensuremath{ \vec{\xfake} }}   %
\newcommand{\xfakebatch}{\ensuremath{ {\bf X'}}}  %
\newcommand{\latent}{\ensuremath{ {\bf z} }}  %
\newcommand{\latentcond}{\ensuremath{ \latent_{\hat c} }}  %
\def\hype{HYPE$_{\infty}$~}

\usepackage[breaklinks=true,bookmarks=false]{hyperref}
\usepackage[capitalize]{cleveref}

\cvprfinalcopy %

\def\cvprPaperID{****} %

\begin{document}

\title{DeepNAG: Deep Non-Adversarial Gesture Generation}

\author{Mehran Maghoumi$^{1,2}$, Eugene M. Taranta II$^2$, Joseph J. LaViola Jr.$^2$\\
	$^1$NVIDIA\\
	$^2$University of Central Florida\\
	{\tt\small mehran@cs.ucf.edu~~~~~etaranta@gmail.com~~~~~jjl@cs.ucf.edu}\\
}

\maketitle

\begin{abstract}
	Synthetic data generation to improve classification performance (data augmentation) is a well-studied problem. Recently, generative adversarial networks (GAN) have shown superior image data augmentation performance, but their suitability in gesture synthesis has received inadequate attention. Further, GANs prohibitively require simultaneous generator \textit{and} discriminator network training. We tackle both issues in this work. We first discuss a novel, device-agnostic GAN model for gesture synthesis called \textnormal{DeepGAN}. Thereafter, we formulate \textnormal{DeepNAG} by introducing a new differentiable loss function based on dynamic time warping and the average Hausdorff distance, which allows us to train DeepGAN's generator without requiring a discriminator. Through evaluations, we compare the utility of DeepGAN and DeepNAG against two alternative techniques for training five recognizers using data augmentation over six datasets. We further investigate the perceived quality of synthesized samples via an Amazon Mechanical Turk user study based on the \hype benchmark. We find that DeepNAG outperforms DeepGAN in accuracy, training time (up to 17$\times$ faster), and realism, thereby opening the door to a new line of research in generator network design and training for gesture synthesis. Our source code is available at \url{https://www.deepnag.com}.
\end{abstract}

\section{Introduction}
\label{sec-intro}

\begin{figure}[t]
	\centering
	\includegraphics[width=0.95\linewidth]{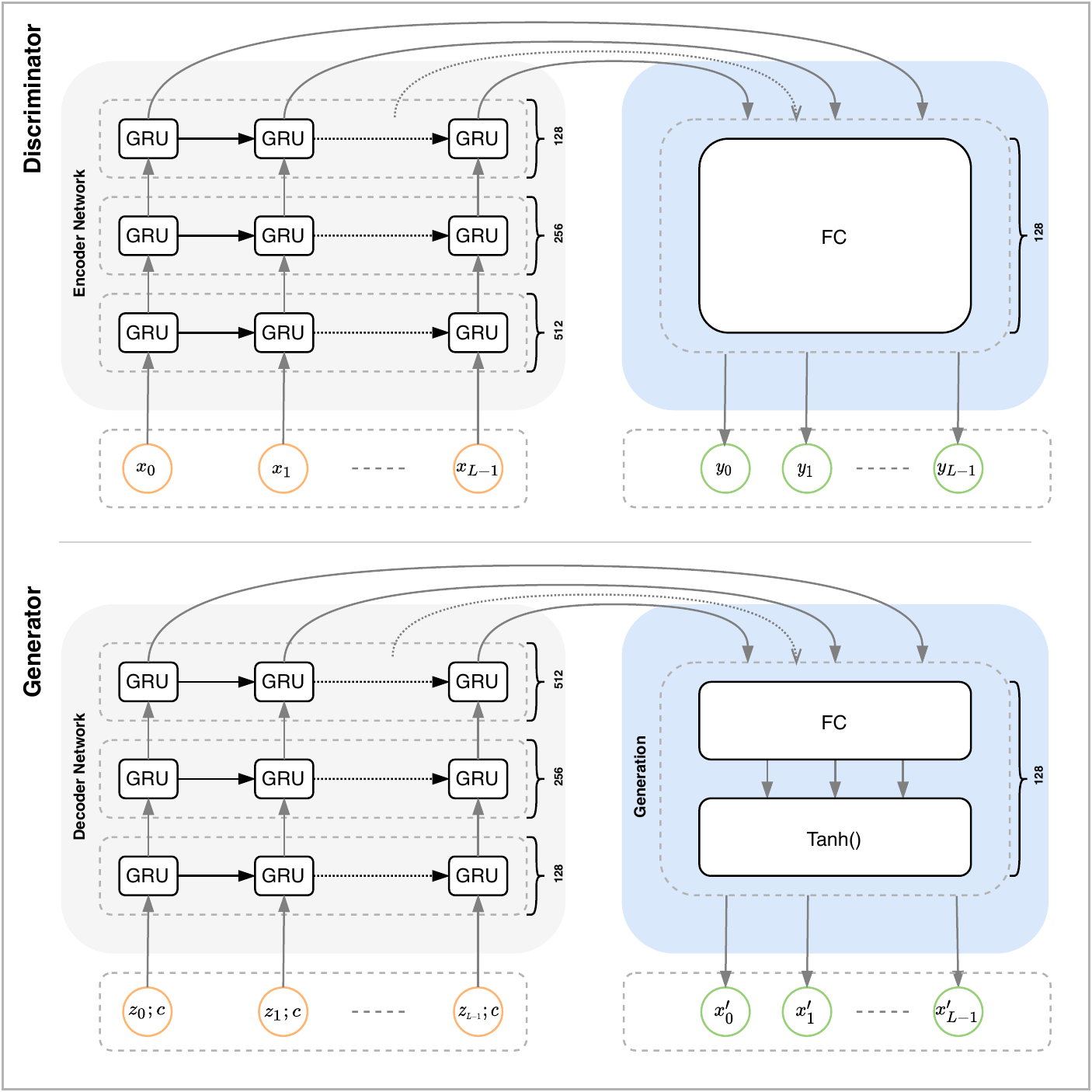}
	\caption{Our proposed model for gesture generation. \textit{DeepGAN} consists of the discriminator and generator networks, whereas \textit{DeepNAG} only consists of the generator network. The generator takes a \textit{class-conditioned} random noise vector as the input and produces a gesture of the specified class. The discriminator (critic) takes the raw gesture points as the input and produces a set of features $y_i$ used in the computation of the Wasserstein loss~\cite{wgan-gp}.}
	\label{fig-gan}
\end{figure}

Recently we observe that system designers are integrating gestures into almost every product with a user interface, igniting the need for accurate gesture recognizers~\cite{part-based-gru,deepgru}. As these recognizers get more sophisticated and accurate, so does their need for more data. While the size of publicly available datasets continues to grow, obtaining more task-specific data is not always easy which highlights the importance of synthetic data generation. Among such methods, generative adversarial networks (GAN)\cite{gan} have shown great promise in various problem domains~\cite{dcgan,rcgan,seqgan}, including handwriting and gesture generation~\cite{bengio-chinese,gesturegan}. Typically, these networks consist of a generator and a discriminator. The discriminator aims to determine if a given example is real or fake, whereas the generator aims to fool the discriminator into confusing fake examples with the real ones. To our knowledge, such models have not received much attention towards \textit{modality-agnostic} gesture generation wherein gestures are represented by a sequence of 2D or 3D positional features typically produced by touch interfaces, Kinect or similar input devices. Additionally, GANs require the concurrent training of two networks, which makes the training procedure long and challenging.

This work focuses on modality-agnostic gesture generation and aims to address the challenges involved with training GANs. We start by discussing \textit{DeepGAN}, our novel GAN approach for dynamic gesture generation, through which our deep recurrent gesture generator network is born. We thereafter discuss our unique solution for alleviating the difficulties associated with training GANs. Specifically, we formulate a novel and intuitive loss function for training our gesture generator. Our loss function, which is based on the dynamic time warping (DTW) algorithm~\cite{dtw}, completely replaces DeepGAN's discriminator network. This, transforms the significantly complex adversarial training procedure of a GAN to the much simpler \textit{non-adversarial} training problem: train the generator by minimizing a loss function that directly maps the quality of the generated examples to their similarity to the real examples. We call this approach \textit{DeepNAG} (see Figure~\ref{fig-gan}). We evaluate both methods by using their generated gestures in data augmentation for improved gesture recognition across a variety of datasets of different sizes and modalities, as well as different gesture recognizers. We additionally conduct a user study to evaluate the human's perception of the realism of our generated results. Such evaluations, which have recently become common practice in the literature of generative modeling~\cite{leiva-large, hype}, provide insight into the visual quality of generated samples.

\vspace{1mm}
\noindent
\textbf{Contributions.} Our main contributions are \textbf{(1)} a novel recurrent GAN model for gesture generation that works across a variety of datasets and modalities, \textbf{(2)} a novel and intuitive loss function that completely replaces the discriminator of our GAN model, which not only simplifies and significantly speeds up the training process, but also yields a generator that produces high quality examples, \textbf{(3)} an evaluation of the improvements in gesture recognition accuracy when our generator is used for data augmentation.

\section{Related Work}
Synthetic data generation is an effective approach in addressing data shortage, which in turn can improve recognition performance~\cite{gpsr,fischer2013historical,varga2005template,lee1998new}. Some data generation methods rely on perturbing existing samples to generate new ones. Taranta~\etal~\cite{gpsr} introduced GPSR which works by selecting random points along a given gesture's trajectory and scaling the between-point distances to create realistic gesture variations.
Other perturbation models include the use of Perlin-noise~\cite{perlin-apply} or the Sigma-Lognormal model~\cite{sigma-lognormal,agogo,kinematic-theory}. These works differ from ours in that we do not rely on inputting existing gestures to produce new ones. Rather, we built a generative model that generates new samples from random noise.

Generative models often involve the use of deep networks. One popular approach that predates GANs is \textit{language modeling}, a probabilistic technique for sequence prediction which works well for handwriting generation~\cite{agraves-generation}. More recently, GANs have gained popularity for such tasks~\cite{rcgan}. Relevant examples include GestureGAN~\cite{gesturegan} a model for hand gesture-to-gesture translation. Given an image of a hand gesture and a target skeleton pose, GestureGAN produces a new hand image holding the target gesture. Yang~\etal~\cite{human-vid-gen} presented a pose-guided human video generation method in which videos of a person performing a desired action are generated. Zhang's~\etal~\cite{bengio-chinese} proposed a recurrent GAN for Chinese character generation which generates the temporal pen movements. The problem domain of these works is different from ours. We focus on modality agnostic gesture generation to produce hand, full-body or 2D pen gestures. Our model learns the representation of a given gesture and produces new gestures in that same representation. Also, our generator can be trained without a discriminator, which sets us apart from the work of Zhang~\etal~\cite{bengio-chinese}.\looseness=-1

Training generative models without a discriminator has also been explored in the literature. Yu~\etal~\cite{seqgan} and Guo~\etal~\cite{leakgan} proposed generating text sequences using reinforcement learning techniques. Lin~\etal~\cite{rankgan} presented the use of a ranking mechanism instead of a discriminator, and Li~\etal~\cite{jsdgan} introduced an adversarial optimization procedure to train a text generator. All of these work focus on generating sequences of discrete tokens (\eg text), whereas our goal is to generate real-valued and continuous multi-dimensional gesture sequences, which is highly challenging as data can take arbitrary values. Lastly, our loss formulation is different from~\cite{jsdgan} in that our formulation is non-adversarial in nature.\looseness=-1

\section{Gesture Generation with Deep Recurrent Networks}
In this section we present our proposed deep learning approaches for gesture synthesis. We first discuss our initial GAN approach from which the DeepNAG generator is born. We then describe the intuition behind our loss function, followed by its formal definition.\looseness=-1

\subsection{Notations and Problem Definition}
In this work, we represent gestures as a temporal sequence of input device samples (\eg 3D joint positions, 2D touch coordinates). At time step $t$, the gesture data is the column vector $x_t \in \realset^{N}$, where $N$ is the dimensionality of the feature vector. Thus, the entire temporal sequence of a single gesture sample is the matrix $\xreal \in \realset^{N \times L}$, where $L$ is the length of the sequence in time steps. For simplicity, and as typical in most gesture recognition work~\cite{jackknife,penny-pincher}, we spatially resample all gesture samples to the same length $L$\footnote{We use $L=64$} as described in~\cite{one_dollar}. We denote the vector trajectory path of a gesture $\xreal$ with $\xrealvec = \{ \overrightarrow{(x_{\scriptscriptstyle i} - x_{\scriptscriptstyle i-1})} ~, ~~ \forall x_{\scriptscriptstyle i > 0} \in \xreal\}$. Lastly, we use $\lvert A \rvert $ to denote the cardinality of a point set $A$, thus $\lvert \xreal \rvert = L$ and $\lvert \xrealvec \rvert = L-1$.

We define gesture generation as producing synthetic examples $\xfake = \{ x_0', x_1', ..., x_{L-1}' \}$ over a dataset of gestures $\dataset$ such these samples mirror data-specific properties of the samples in $\dataset$, as if these examples were seemingly sampled from $\dataset$. Formally, if $p_{\dataset}$ is dataset's distribution such that $\xreal \in \dataset \implies \xreal \sim p_{\dataset}$, our goal is to synthesize $\xfake$ where $\xfake \notin \dataset$ but $\xfake \sim p_\dataset$. We aim to achieve this using a deep recurrent network $G$ which maps a \textit{class-conditioned} latent vector $\latentcond$ to a synthetic example $\xfake = G(\latentcond; \theta_G)$, where $\theta_G$ are the trainable parameters of $G$. Henceforth, we use $G_\theta(\latentcond)$ in place of $G(\latentcond; \theta_G)$, and use $\loss$ to denote an objective function to minimize (a training loss function).

\subsection{Gesture Generation with GANs}
\label{sec-gan}
Our initial approach for gesture generation uses the well-known GAN training setting comprised of a generator and a discriminator. We call this recurrent model \textit{DeepGAN}, which we designed incrementally and was informed by the latest developments in deep learning. Early on, the simplicity and the recognition power of the recently proposed DeepGRU model~\cite{deepgru} inspired us to adopt it as our discriminator. This encoder-style model has shown promising results in various recognition tasks~\cite{udeepgru,part-based-gru}. Through experiments guided by an ablation study on DeepGRU~\cite{deepgru}, and with the goal of managing design complexity, we settled for the simpler \textit{uDeepGRU}~\cite{udeepgru} variant as our discriminator. As for our generator, we conducted experiments across different datasets with generators consisting of both LSTM and GRU units, as well as a varying number of recurrent layers. We observed stabler training, less overfitting and more plausible outputs with a decoder-style network resembling the \textit{flipped} version of our discriminator. A possible explanation for this could be that this choice potentially benefits from the balance between the two $D$ and $G$ networks. Figure~\ref{fig-gan} depicts the architecture of DeepGAN, which we believe is easy to understand and straightforward to implement in any modern deep learning framework. A common design for generators is the use of the $tanh()$ activation function in the last layer, which is what we use as well.

To generate a gesture sample $\xfake$ of class $c$, the class-conditioned latent vector $\latent_{\hat c}$ is fed to $G$ where $\latentcond$ is defined as Equation~\ref{eq-class-conditioned}. Note that each time step $z_i \in \latent_{\hat c}$ is sampled independently from the standard normal distribution\footnote{The dimensionality of the latent space was fixed to 32 dimensions.}, and class-conditioning is done by appending the \textit{one-hot} representation of $c$ to each time step which avoids ignoring the conditioning through forgetting~\cite{rcgan}.

\begin{align}
\label{eq-class-conditioned}
\latentcond &= \Big\{ [~z_i; \hat{\bf c}~]:~~\forall z_i \sim \mathcal{N}\big(0, 1\big),~ \hat{{\bf c}} = \text{one-hot}\big( c \big)  \Big\}
\end{align}

We experimented with different loss functions to train DeepGAN. Even though training with the classic adversarial loss~\cite{gan} yielded plausible results, we observed improved sample quality and better convergence with the improved version~\cite{wgan-gp} of the Wasserstein loss (WGAN)~\cite{wgan}, which is what we settled on using for our evaluations. Figure~\ref{fig-gan} depicts DeepGAN's architecture.

\subsection{Non-Adversarial Gesture Generation}
\label{sec-nag}
Although DeepGAN shows promising results (see Section~\ref{sec-evaluation}), it demonstrated a few shortcomings early on. Most importantly, the need for training two networks simultaneously increases the training burden: changes in one network may adversely affect the other and most hyperparameters need to be tuned twice. Moreover, training times are long and we observed that the model required tens of thousands of generator iterations to converge. Aiming to reconcile these challenges, we present our loss function that completely replaces the discriminator. We start by providing an intuition for training a sequence generator without a discriminator, then proceed with the formal definition of our loss function.

\vspace{1mm}
\noindent
\textbf{Intuition.~}
The key to simplify the network design is the answer to a fundamental question: can we possibly train the single network generator network $G$? A generator network aims to learn the distribution of the underlying dataset \dataset, so that new examples can be sampled from the distribution, which is typically done with the help of a discriminator (or a critic, in the case of WGAN-based models). 
To simplify the gesture generation procedure, we pose the problem in a slightly different way: let us train a generator that aims to produce gestures that are \textit{similar} to their real counterparts. The fundamental question then becomes, how to train a generator that increases the similarity between the generated and the real gestures? The answer is surprisingly simple: by reducing the \textit{dissimilarity} between the two! Conveniently, a well-studied dissimilarity metric for time-series (as well as gestures~\cite{jackknife}) is \textit{dynamic time warping} (DTW)~\cite{dtw}. A differentiable formulation of DTW called soft DTW (sDTW) was recently proposed by Cuturi and Blondel~\cite{sdtw}.

DTW is a dynamic programming algorithm that was originally proposed for speech recognition~\cite{dtw}. It is a dissimilarity measure of two time-series that can be used to find their optimal alignment for various time-series analysis tasks~\cite{dtw-trillion}. Given two time-series $\X = \{ x_1, x_2, ..., x_n \}$ and $\Y = \{ y_1, y_2, ..., y_m  \}$, a cost matrix $\Delta$ of size $n\times m$ is built. Each element $\Delta_{ij}$ is the matching cost of $x_i$ to $y_j$, computed via the following recursion:

\begin{equation}
\label{chap-background-eq-dtw}
\Delta_{ij} = f\big( x_i, y_j \big) + \text{min}\bigg\{
\Delta_{i-1, j},
\Delta_{i, j-1},
\Delta_{i-1, j-1}\bigg\}
\end{equation}

\noindent
where $f\big( x_i, y_j \big)$ is a problem-specific cost (distance) function. Although Euclidean distance (ED) is widely used, Taranta~\etal~\cite{jackknife,penny-pincher} demonstrated the superiority of using the cosine similarity metric (COS) for gesture recognition problems. Once $\Delta$ is fully computed, the value $\Delta_{nm}$ is the dissimilarity measure of $\X$ and $\Y$, and the path through the matrix that yields $\Delta_{nm}$ is the optimal alignment between the two time-series. Cuturi and Blondel's sDTW formulation~\cite{sdtw} replaces the $\text{min}\{\}$ operator with a differentiable (soft) minimum defined as:

\begin{equation}
\label{chap-background-eq-softmin}
\text{min}^{\scriptscriptstyle \gamma > 0}\bigg\{ \Delta_1, \Delta_2, ..., \Delta_n \bigg\} = 
-\gamma~\text{log}\sum\limits_{i=1}^{n}{e^{-\Delta_i / \gamma}}
\end{equation}

\noindent
where $\gamma$ controls the smoothness (smaller $\gamma$ yields a closer approximation of classic DTW). Cuturi and Blondel show that the resulting $\Delta_{nm}$ would be the expected value of dissimilarity between $\X$ and $\Y$, over every possible alignment between them weighted by their probability under the Gibbs distribution~\cite{sdtw}. Further, they formulate the derivative of sDTW using backpropagation through a computation graph. A detailed explanation is available in ~\cite{sdtw}.

\vspace{1mm}
\noindent
\textbf{Loss formulation.~}
To increase the similarity of a real example $\xreal$ and a fake example $\xfake$, one could na\"ively decide to learn $\theta$ by minimizing $\loss = \text{sDTW}\big(G_\theta(\latentcond), {\xreal}; f)$ where $f$ is the cost function of Equation~\ref{chap-background-eq-dtw}. Unfortunately, this formulation merely states that generated samples $\xfake$ should be as similar to the real samples $\xreal$ as possible, ignoring inter-class variations. A trivial solution to this formulation, that was easily achievable in our tests, is the per-class centroid sample. To overcome this issue and account for inter-class variability, we formulate the loss function as a coverage measure (point set similarity) of a set of fake and real examples. For this, we propose using the \textit{Hausdorff distance}, a well-studied measure for point set similarity~\cite{mhd} that can be easily implemented and computed. Conveniently, the average Hausdorff distance (denoted as $\hd$) between two point sets $\bf A$ and $\bf B$ is differentiable~\cite{ahd-diff} and is defined as:

\begin{align} 
\label{chap-deepnag-eq-hd}
\hd^f({ \bf A}, {\bf B}) ~=~ 
&\frac{~~1~~}{\lvert \bf A \rvert} \mathlarger{\sum}\limits_{a \in {\bf A}} \operatorname*{min}_{b \in {\bf B}}~d\Big( a, b; f \Big) ~+~\\
&\frac{~~1~~}{\lvert \bf B \rvert} \mathlarger{\sum}\limits_{b \in {\bf B}} \operatorname*{min}_{a \in {\bf A}}~d\Big( b, a; f \Big)\nonumber
\end{align}

\noindent
where $d\big(a, b; f\big)$ is the distance (dissimilarity) between two points $a$ and $b$ parameterized by $f$. We use $d\big(a, b; f\big) = \text{sDTW}(a, b; f)$, and will discuss the choice of $f$ (sDTW's cost function) shortly. Finally, we propose the following as DeepNAG's loss function to minimize. To our knowledge, this is the very first formulation of a single loss metric to train a deep recurrent neural network for generating synthetic gesture sequences:

\begin{gather}
\label{chap-deepnag-eq-loss}
\loss_{\tiny f}(\xfakebatch, \xrealbatch) = 
\underbrace{
	\mystrut{2ex}
	\hd^f\big( \xfake_1, \xreal_1 \big)
}_{\text{Similarity term}}
+
\underbrace{
	\mystrut{2ex}
	\Big\lvert
	\hd^f\big( \xfake_1, \xfake_2 \big) - 
	\hd^f\big( \xreal_1, \xreal_2 \big)
	\Big\rvert
}_{\text{Variation term}}
\end{gather}

\noindent
where $\xfakebatch=\{\xfake_1$, $\xfake_2\}$ (two generated examples), and $\xrealbatch = \{\xreal_1$, $\xreal_2\}$ (two real examples), and both examples sets belong to the same gesture class. We only use the derivative ${\partial \loss_{\tiny f}} / {\partial \xfake_1}$ during training since it yielded good results and was faster. Intuitively, Equation~\ref{chap-deepnag-eq-loss} expresses that training $G$ should aim to increase the similarity of fake and real examples\footnote{In other words \textit{decrease} their \textit{dissimilarity}} (similarity term), while maintaining the similarity balance between two batches of fake and real examples (variation term).
The former term ensures real and generated samples are similar, while the latter term ensures generated examples maintain proper overall inter-class variations, effectively avoiding pitfalls such as mode-collapse typically encountered in GANs.\looseness=-1

\vspace{1mm}
\noindent
\textbf{Practical notes.~} When computing $\text{sDTW}(a, b)$ we ensure \textit{class-awareness}: the value is only computed if samples $a$ and $b$ belong to the same gesture class. As for the choice of sDTW's internal cost function $f$, we started with ED, but the benefits of using COS quickly became apparent to us: minimizing $\loss_{\text{\tiny ED}}$ yielded high-quality results but convergence was slow. Conversely, minimizing $\loss_{\text{\tiny COS}}$ led to much faster convergence with sometimes noisier results. In the end, we settled for minimizing both and leave a thorough study on the effects of each cost function to future work. Lastly, recall our use of fixed-length sequences ($L=64$) where the points in the sequence are equidistant. To enforce the production of such sequences by $G$ we add an additional term $\loss_{\text{\tiny Resample}}$ to our objective. Putting everything together, the following is the loss function that we minimize for our experiments:\looseness=-1

\begin{align}
\label{full-nag-loss}
\loss_{\text{\tiny DeepNAG}}\big(\xfakebatch, \xrealbatch\big) =~& 
\loss_{\text{\tiny ED}}\big(\xfakebatch, \xrealbatch\big)
+
\loss_{\text{\tiny COS}}\big(\xfakebatch, \xrealbatch\big)
+\\
&{\tiny \alpha} \cdot \loss_{\text{\tiny Resample}}\big(\xfake_1\big)\nonumber
\\
\loss_{\text{\tiny Resample}}\big(\xfake\big) =~& 
\frac{1}{\lvert \xfakevec \rvert}
\mathlarger{\sum}\limits_{\scriptscriptstyle \forall \vec{x_{\scriptscriptstyle i}'} \in \xfakevec}
{ 
	\bigg(
	\norm{\vec{x_i'}} - \widetilde{L}(\xfake)
	\bigg)^{\scriptscriptstyle 2}
}\nonumber
\\
\widetilde{L}(\xreal) =~& \frac{~~1~~}{\lvert \xrealvec \rvert} \mathlarger{\sum}\limits_{\scriptscriptstyle \forall \vec{x_{\scriptscriptstyle i}} \in \vec{\xreal}}{~\norm{~ \vec{x_{\scriptscriptstyle i}}~ }}\nonumber
\end{align}

\noindent
where $\widetilde{L}(\xreal)$ is the length of each $\vec{x_i} \in \xrealvec$ after $\xreal$ is resampled to $L$ equidistant points.
Thus, $\loss_{\text{\tiny Resample}}$ simply enforces that points in $\xfake$ be equidistant with $\alpha$ as its regularizer. Note that minimizing $\loss_{\text{\tiny ED}}$ alone (Equation~\ref{chap-deepnag-eq-loss}) yields good-quality results in most cases. However, we achieved faster convergence and better data augmentation performance using Equation~\ref{full-nag-loss}.

As we discuss shortly, a generator trained with our loss function demonstrates promising results compared to when the same generator is trained in a GAN setting. In addition to these two training configurations, we considered training our generator in a variational autoencoder (VAE)~\cite{vae} setting. To our knowledge, no prior work has discussed the adaptation of a recurrent VAE for gesture generation. Although we successfully generated gesture sequences using our proposed VAE, the quality and the variety of our produced samples were sub-par compared to either DeepGAN or DeepNAG. Refer to Appendix~\ref{appendix} for more details.\looseness=-1

\section{Evaluation}
\label{sec-evaluation}
We evaluate DeepGAN and DeepNAG from two aspects. First, we conduct experiments to determine the efficacy of either model in data augmentation tasks, focusing on scenarios with limited training data. We then discuss the evaluation of the perceived realism of our synthetic gestures through a user study on Amazon Mechanical Turk based on a recently introduced benchmark~\cite{hype}.

\subsection{Data Augmentation Performance}
\label{sec-evaluation-aug}
Our experiment design for this study is as follows. Given a dataset of gestures collected from multiple participants, we simulate small training sets by splitting the data into training (50\%), validation (20\%) and test (30\%) sets. Our experiments are all \textit{subject-independent}; \ie the data of each participant only appears in one of these sets. This is a more challenging and realistic evaluation protocol, as it ensures that during training, the recognizer never sees any data from the participant that it will be evaluated on during the validation and testing phases.

\begin{figure*}[tbh]
	\ra{0.5}
	\centering
	\begin{tabular}{c c}
		\includegraphics[width=0.45\linewidth]{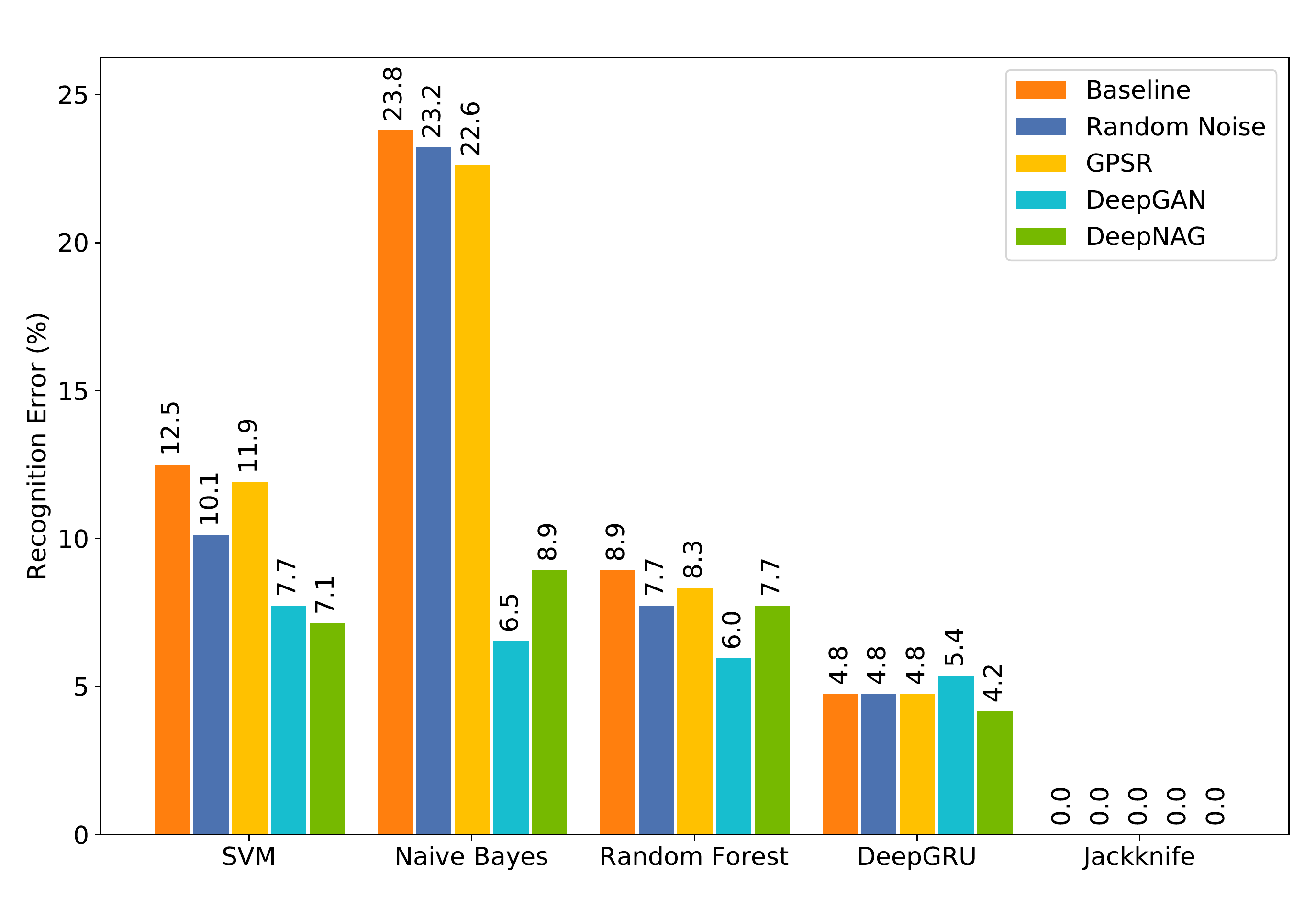} &
		\includegraphics[width=0.45\linewidth]{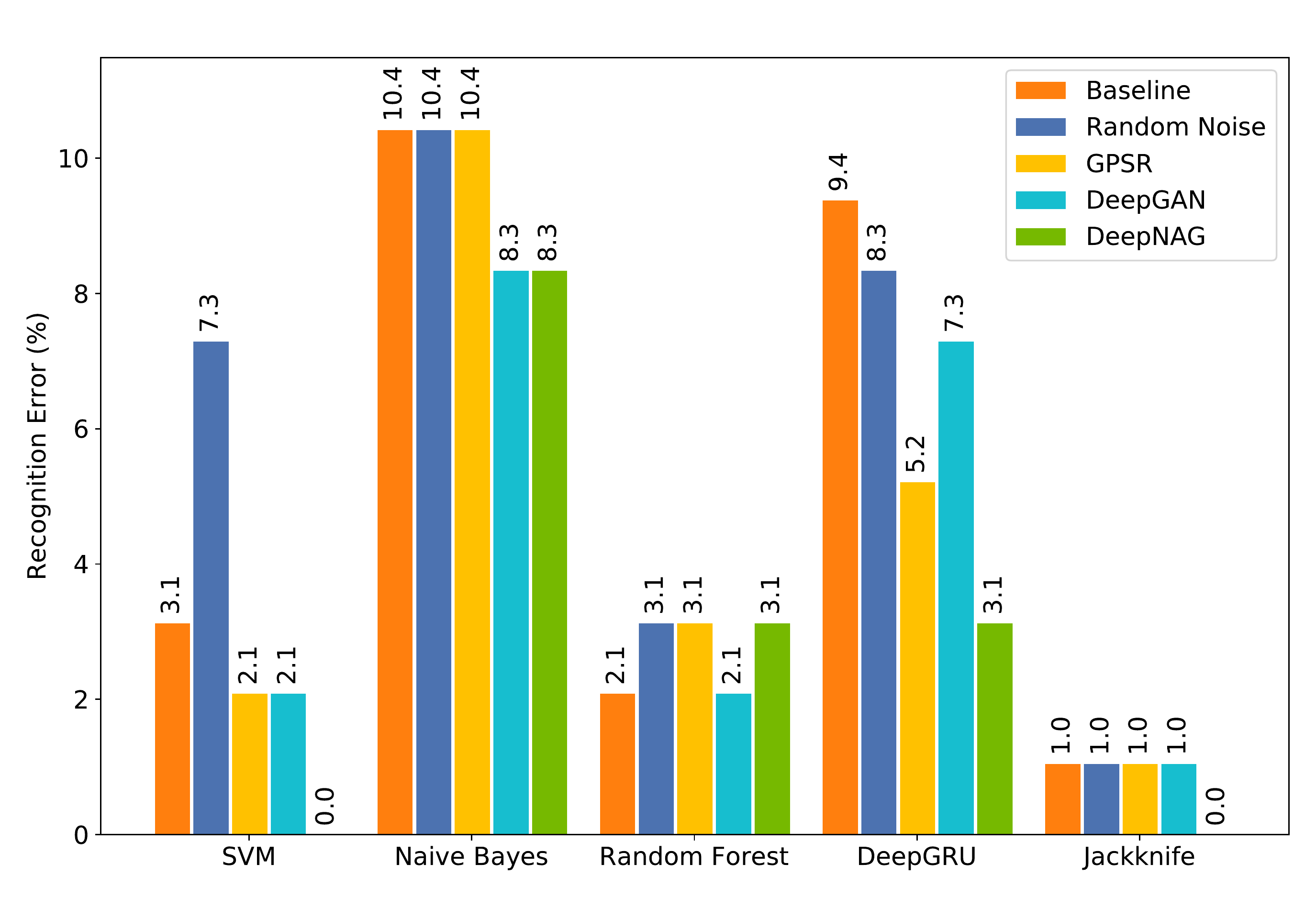}\\
		(a) JK2017 (Kinect)~\cite{jackknife} & (b) JK2017 (Leap Motion)~\cite{jackknife}\\
		\includegraphics[width=0.45\linewidth]{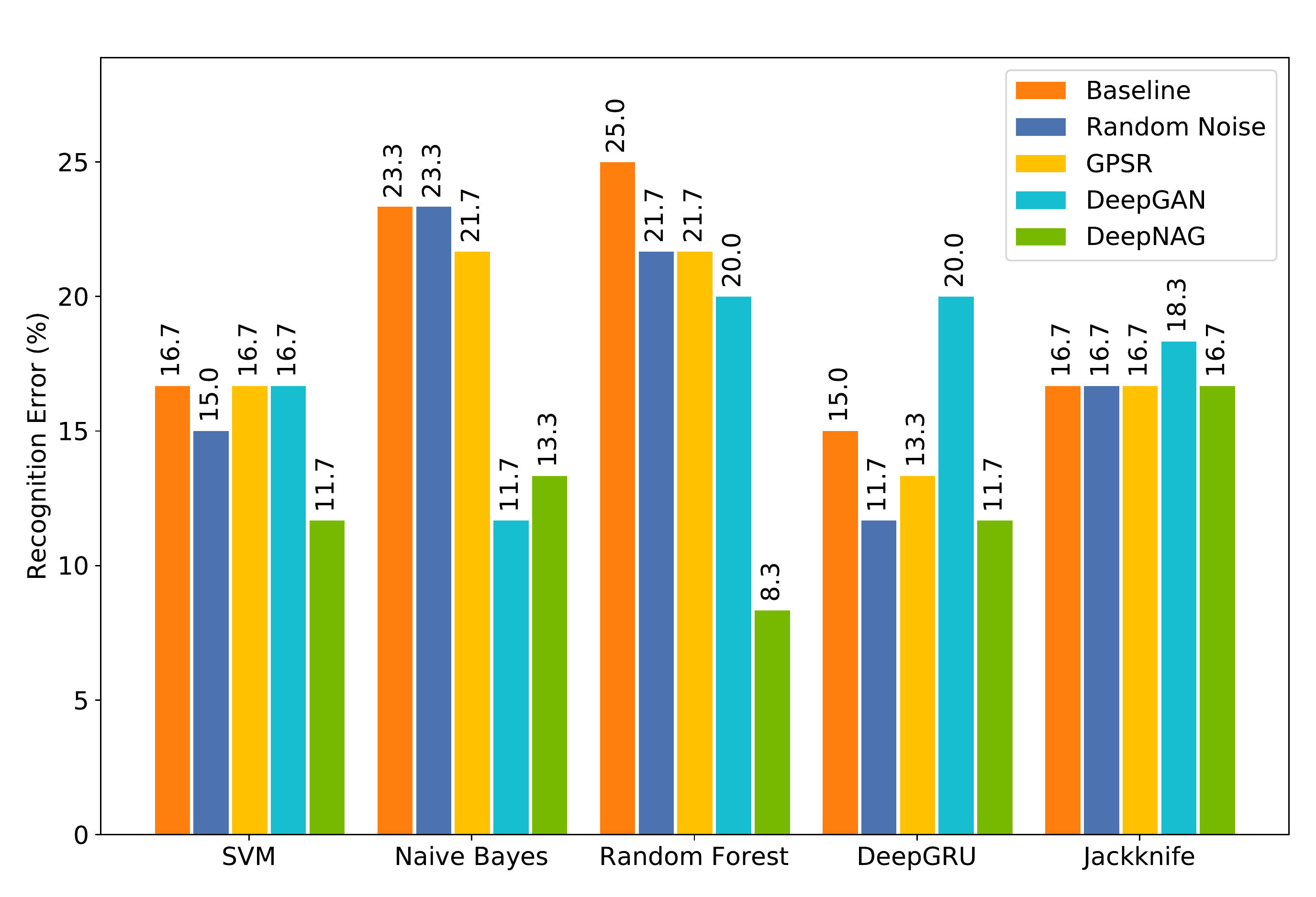} &
		\includegraphics[width=0.45\linewidth]{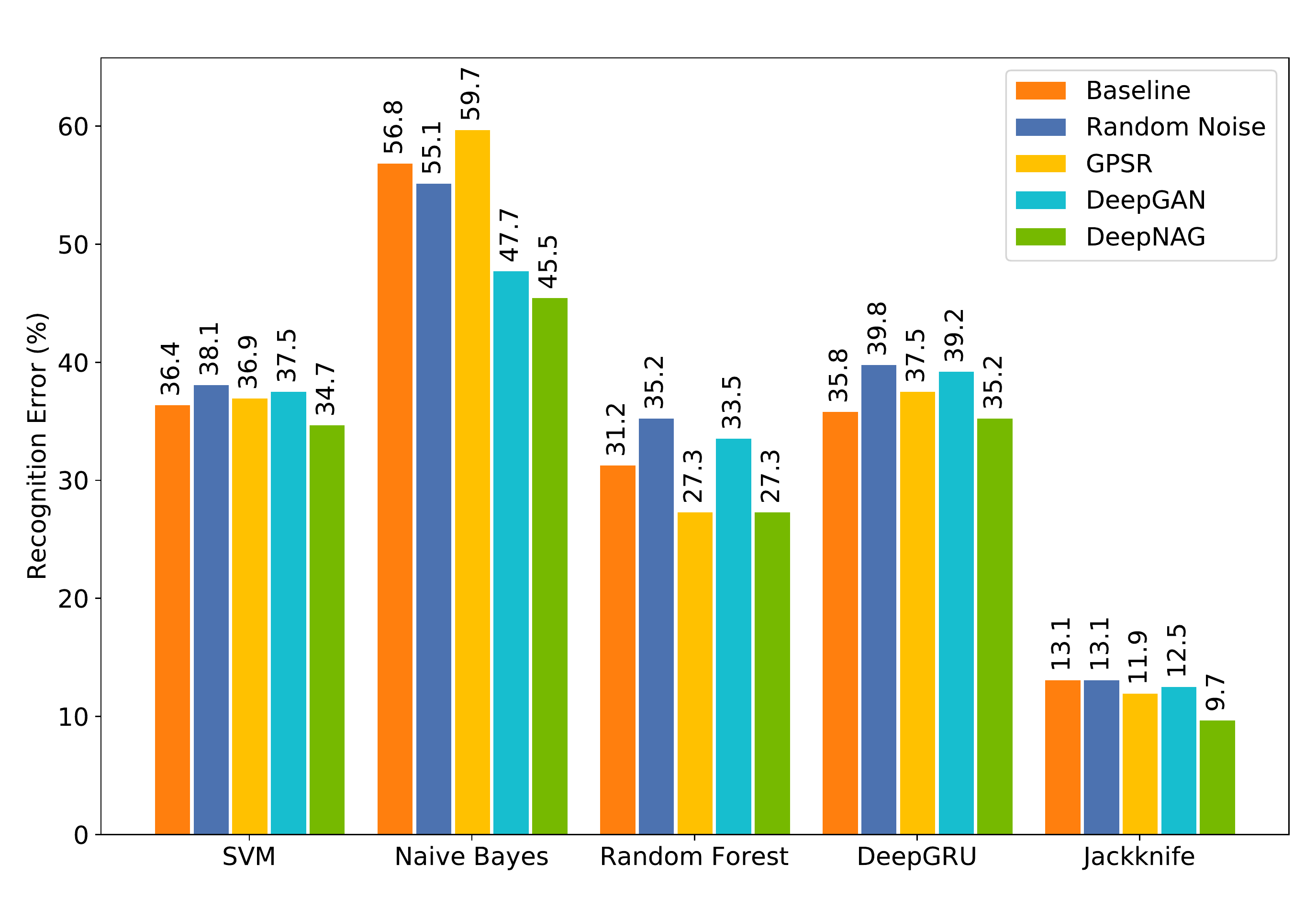}\\
		(c) UT-Kinect~\cite{ut-kinect} & (d) MSR Action3D~\cite{msr-action3d}\\
		\includegraphics[width=0.45\linewidth]{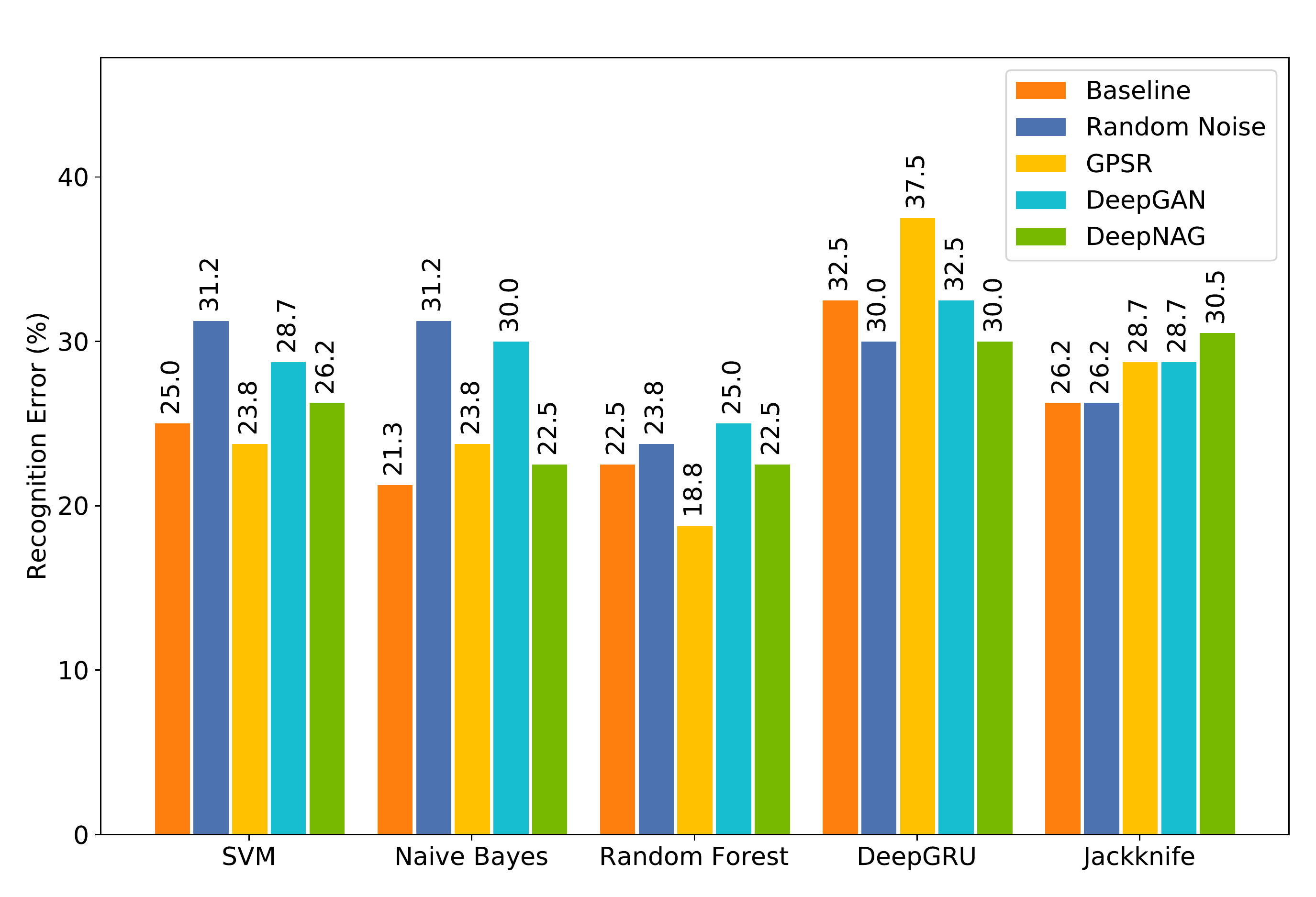}
		&
		\includegraphics[width=0.45\linewidth]{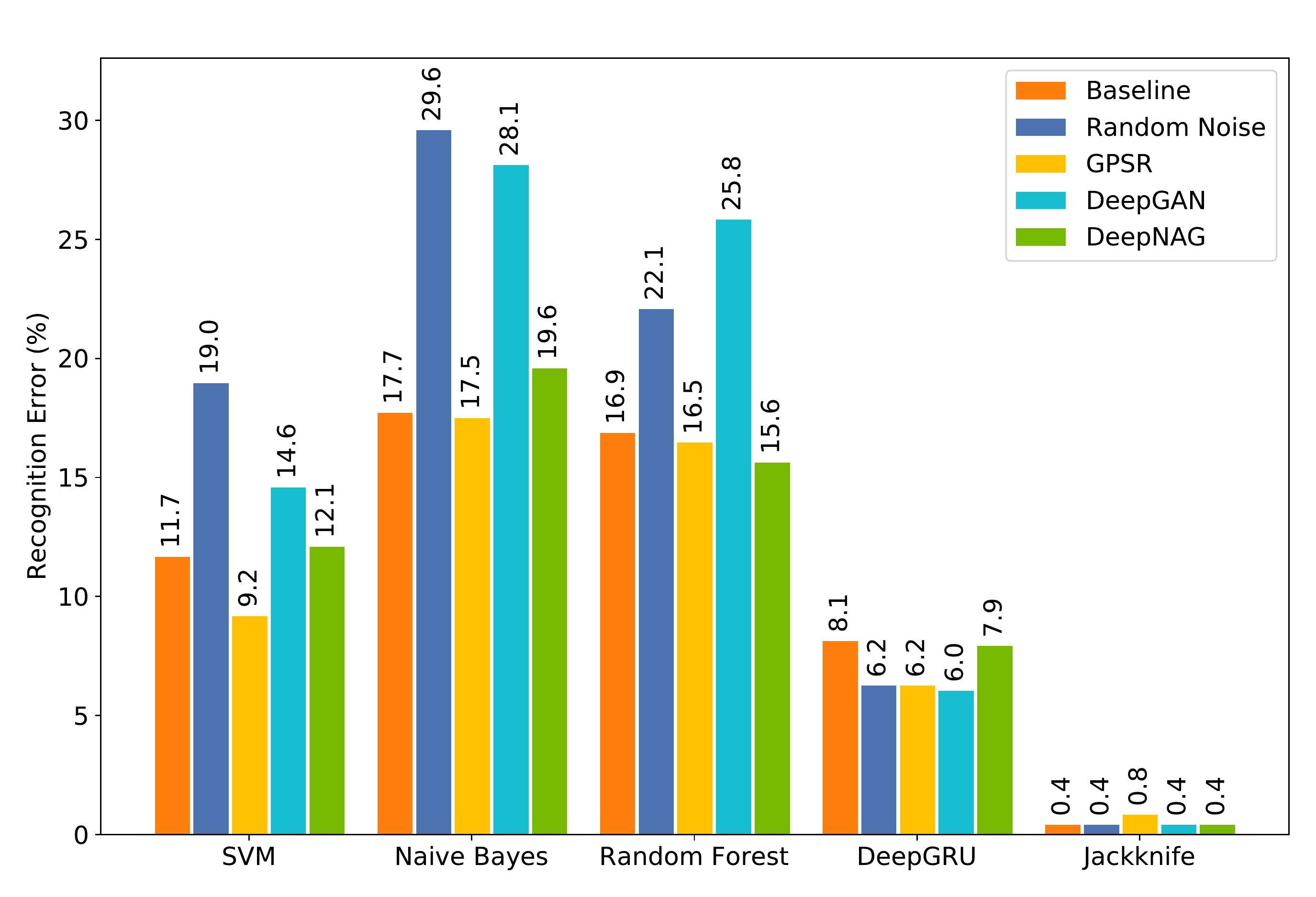}\\
		(e) SBU Kinect Interactions~\cite{sbu-kinect} & (f) \$1-GDS~\cite{one_dollar} \\
	\end{tabular}
	\caption{Results of evaluation across six datasets (best viewed in color).}
	\label{fig-deepgan-results}
\end{figure*}

We begin by training a gesture recognizer on the training set, and use the validation set for model selection. We evaluate the best performing model on the test set and record its recognition error (baseline). Next, we augment the training set with a selected data generation method and repeat the experiment: train the recognizer with this new training set, use the validation data for model selection, and evaluate the best model on the test set. We record the recognizer's recognition error again, which will be the error after augmenting the training set. Comparing this number with the baseline benchmarks the synthetic data generation method. In total we perform 150 experiments: we train five gesture recognizers on six different datasets to evaluate four synthetic data generation methods against the baseline.

\vspace{1mm}
\noindent
\textbf{Datasets.} We selected six datasets among the ones frequently studied in the literature. They vary in size and span across gesture modalities and input devices: JK2017 (Kinect)~\cite{jackknife} (14 full-body fighting gestures of 20 participants with Kinect v2), JK2017 (Leap Motion)~\cite{jackknife} (eight hand-gestures of 20 participants with Leap Motion), UT-Kinect~\cite{ut-kinect} (ten full-body daily activities of ten participants with Kinect v1), MSR Action 3D~\cite{msr-action3d} (20 full-body actions of ten participants with Kinect v1), SBU Kinect Interactions~\cite{sbu-kinect} (8 two-person interaction of seven participants with Kinect v1) and \$1-GDS~\cite{one_dollar} (16 2D pen gestures of ten participants).

\begin{table*}[tbh]
	\small
	\centering
	\ra{1.2}
	\begin{tabular}{l c c c l l c c c}
		\toprule
		\multirow{2}{*}{\textbf{Dataset}} & \multicolumn{3}{c}{\textbf{Generator Score}} && \multirow{2}{*}{\textbf{Recognizer}} & \multicolumn{3}{c}{\textbf{Generator Score}}\\
		\cmidrule(l{4pt}r{4pt}){2-4} \cmidrule(l{4pt}r{4pt}){7-9}
		& {\small \textbf{GPSR}} & {\small \textbf{DeepGAN}} & {\small \textbf{DeepNAG}} &&& {\small \textbf{GPSR}} & {\small \textbf{DeepGAN}} & {\small \textbf{DeepNAG}}\\
		\midrule
		JK2017 (Kinect)~\cite{jackknife} & 0 & 2 & 2
		&&
		SVM & 2 & 0 & 4\\
		
		JK2017 (LeapMotion)~\cite{jackknife} & 0 & 1 & 4 
		&&
		Na\"ive Bayes & 1 & 3 & 2\\
		
		UT-Kinect~\cite{ut-kinect} & 0 & 1 & 2
		&&
		Random Forest & 2 & 1 & 3\\

		MSR Action3D~\cite{msr-action3d} & 1 & 0 & 3
		&&
		DeepGRU~\cite{deepgru} & 0 & 1 & 3\\
		
		SBU Kinect & 2 & 0 & 0
		&&
		Jackknife~\cite{jackknife} & 0 & 0 & 2\\
		
		\$1-GDS~\cite{one_dollar} & 2 & 1 & 1\\
		\midrule
		\textbf{Total Score} & 5 & 5 & \textbf{12}
		&&&
		5 & 5 & \textbf{14}\\
		\bottomrule
	\end{tabular}
	\caption{Generator scores aggregated over \textit{dataset} and \textit{recognizer}.}
	\label{tab-scores}
\end{table*}

\vspace{1mm}
\noindent
\textbf{Recognizers.} We selected five gesture recognizers: support vector machine (SVM), random forest, na\"ive Bayes, DeepGRU~\cite{deepgru} and Jackknife~\cite{jackknife}. These represent classic machine learning algorithms, deep learning as well as rapid prototyping~\cite{jackknife} approaches, which are common choices for gesture recognizers. The first three methods require explicit feature extraction for which we use the Rubine~\cite{rubine} feature set extended to 3D gestures~\cite{3d-rubine}. Jackknife~\cite{jackknife} is a 1-nearest neighbor DTW-based template matching recognizer.

\vspace{1mm}
\noindent
\textbf{Data generation methods.} We compare four data generation methods against the baseline: random Gaussian noises, GPSR~\cite{gpsr}, DeepGAN and DeepNAG. Although GPSR was originally used for 2D gestures, its effectiveness for 3D gestures has been demonstrated~\cite{deepgru,udeepgru}.

\vspace{1mm}
\noindent
\textbf{Implementation.} We implemented DeepGAN and DeepNAG with the PyTorch~\cite{pytorch} framework which we have publicly released. Additionally, our implementation requirements yielded multiple other standalone projects, which we have made public in the hope of benefiting the deep learning community. Inspired by~\cite{dtw-diagonal}, we implemented a CUDA version of sDTW with a PyTorch interface using Numba~\cite{numba}. Our novel implementation parallelizes both forward and backward passes, and runs more than $100\times$ faster than any other publicly available implementation that we know of. Additionally, we implemented fast GRU units using PyTorch's just-in-time (JIT) compilation features to allow computing their higher-order derivatives, a feature that is missing in PyTorch\footnote{To our knowledge, the cuDNN framework is missing this feature. Thus, at the time of this writing one cannot compute higher-order derivatives for GPU-based GRUs in any deep learning framework that relies on cuDNN.}. Such derivatives are required to implement the improved WGAN loss~\cite{wgan-gp} for GRUs.

\vspace{1mm}
\noindent
\textbf{Hyperparameters.~} All hyperparameters were tuned across different datasets, but the same set of parameters were used for every experiment. Both DeepGAN and DeepNAG were trained on the 50\% split training set, and shared most hyperparameter settings. We use the Adam~\cite{adam} solver ($\beta_1=0.5, \beta_2=0.9$), with a learning rate of 10$^{-4}$ and a mini-batch size of 64. DeepGAN-specific hyperparameters were chosen from~\cite{wgan-gp}, as they performed the best in our validation runs. Other parameters were chosen via cross-validation as follows. For DeepNAG we used $\gamma=0.1$ and $\alpha=10^{3}$. GPSR parameters were set to $r=2, \sigma=0.25$ and the magnitude of random noise was set to 2\% of the bounding box of each feature.\looseness=-1

\vspace{1mm}
\noindent
\textbf{Results and discussion.~}
Figure~\ref{fig-deepgan-results} depicts the results of our experiments\footnote{A video demo of generated gestures is available in an accompanying video. Visit \url{https://www.deepnag.com}}. In many cases the use of some form of data augmentation decreases the recognition error, indicating that our 50\% split to simulate small training sets is working as expected. To better contrast the generation methods we employ a scoring scheme that quantifies whether the use of a given augmentation method is both \textit{warranted} and \textit{effective}. Data augmentation is only warranted if a recognizer trained with the additional data outperforms the baseline. Additionally, a method is effective only if it outperforms random noise. We start with a score of zero for a given generator. In each experiment set, we increment this score by one if the method outperforms all other methods in addition to the baseline and random noise. Ties are only counted if the method outperforms both random noise and the baseline, and we use the cumulative score for comparison.

Table~\ref{tab-scores} presents the computed score aggregates over each dataset and recognizer. We observe that across both aggregate groups, DeepNAG outperforms other methods by a large margin, suggesting its suitability for data augmentation regardless of the choice of dataset or recognizer. In a few cases, DeepNAG reduced the recognition error to zero, which further supports its suitability. Compared to GPSR, these results are notable as DeepNAG generates new examples purely from random noise. Conversely, GPSR perturbs existing examples to generate new ones. This process leaves some characteristics of the original gesture (\eg bounding box size) largely unchanged, which benefits recognizers that rely on such features.

Figure~\ref{fig-deepgan-results} also shows cases wherein data augmentation seems harmful. In particular, we observe increased errors in almost all cases where data generation is used with multi-actor gestures (Figure~\ref{fig-deepgan-results}e). This suggests that our generators may not be suitable for generating multi-actor gestures, which we confirmed by visual inspection. In some cases, both DeepGAN and DeepNAG confuse the main and the secondary actors, yielding malformed gestures. We intend to study the generation of such gestures in future work. We additionally inspected some of the generated examples of Figure~\ref{fig-deepgan-results}f wherein our generators increased recognition errors. Most synthetic examples were visually fine which suggests that the use of domain adaptation techniques may be helpful~\cite{cycada,cyclegan,gan-image-classification}. We plan to explore this in the future.\looseness=-1

During visual inspection, we did not observe any mode-collapse issues with DeepNAG. We observed healthy variations across all gesture classes and datasets with minimal amounts of degenerate samples (except for the few cases noted above). Lastly, factors such as ease of training and training times compel the use of DeepNAG over DeepGAN as the former offers a significant reduction in training times. Training DeepGAN on a Tesla V100 GPU takes between 3-5 days depending on the dataset size, whereas DeepNAG takes around 3-7 \textit{hours} under the same conditions, a speedup of 12--17$\times$.

Overall, our results indicate that DeepNAG outperforms DeepGAN on data augmentation tasks, regardless of the choice of dataset or recognizer. These results are notable, as the generator model in both DeepGAN and DeepNAG is exactly the same. In other words, the generator which is trained using our novel loss function outperforms the same generator trained in a GAN setting with the improved Wasserstein loss. Additionally, our loss function trains the generator in a much shorter period of time.

\subsection{User Study}
Qualitative evaluation of generative models through user studies has become a common practice in the literature~\cite{leiva-large, hype}. We therefore turn our focus to the comparison of DeepGAN and DeepNAG based on the perceived quality of the generated samples using human evaluators based on the \hype~\cite{hype} benchmark. This benchmark defines the gold standard for evaluating generative realism on crowd-sourcing platforms.

To compare different generative models, \hype defines an experiment with 30 participants: each participant only sees the results from one of the models. For a given generative model, every participant is shown a total of 100 samples comprised of 50 fake and 50 real samples. Given each sample, participants are asked to indicate whether they think that sample is real or computer-generated. Participants have an infinite amount of time to make this binary choice. Afterwards, the percentage of the samples that were judged incorrectly is computed for every participant. Obtained values are averages over 30 participants and the final result is reported as the \hype score for the understudied generative model. Zhou~\etal~\cite{hype} showed that this protocol ensures repeatability and maintains the separability between different generative models and can be used as a reliable measure of the generative model's quality. 

\begin{figure}[t]
	\centering
	\includegraphics[width=0.9\linewidth]{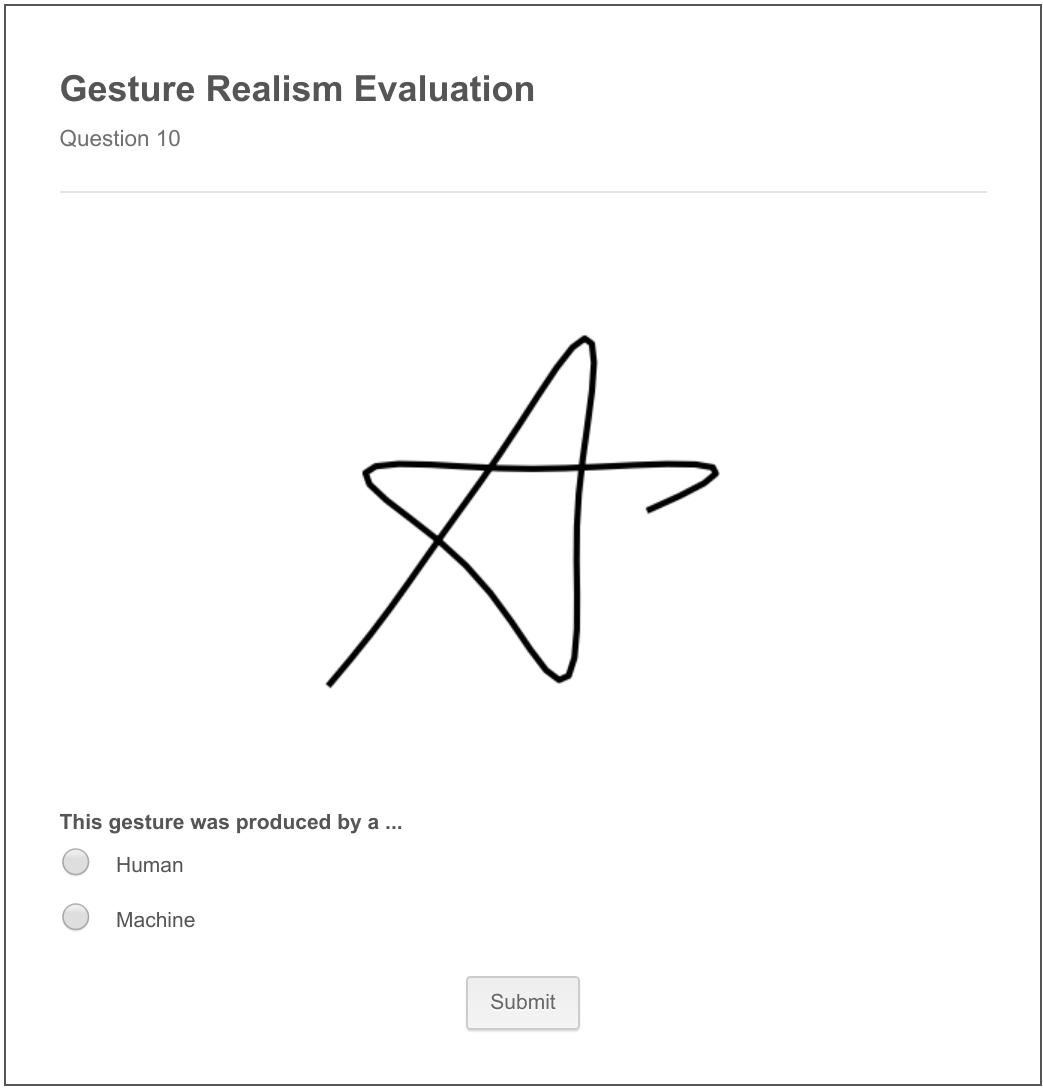}
	\caption[User study UI]{The interface of our user study application. Participants are shown the gesture animation and are asked to select either ``human" or ``machine". Once ``submit" is clicked, the correct answer is revealed.}
	\label{chap-deepnag-fig-study}
\end{figure}

Using this protocol, we conduct our user study on Amazon Mechanical Turk for a given dataset $\dataset$ and a generator $G$. We first train the $G$ on $\dataset$ to convergence. Using the trained model, we then sample as many fake samples as the real samples in $\dataset$. The 100 samples needed to show a given participant are randomly drawn from the pool of all available samples. We recruit 30 participants for every combination of $\dataset$ and $G$. Participants begin by studying the purpose of the study and answering demographic questions as detailed in Table~\ref{chap-deepnag-tab-prequestions}. We then randomly show them each of the 100 samples and ask them to indicate whether they think a given sample is produced by human or computer-generated (see Figure~\ref{chap-deepnag-fig-study}). Every gesture sequence is drawn in the form of a looping \textit{gif} animation with a framerate of 32. Between each animation loop, we display a countdown with a duration of 0.25 seconds. This was inspired by~\cite{hype} and was done to avoid confusing participants who may be unaware that they are watching an infinitely-looping animation.

Participants are given an infinite amount of time to respond to each question. Similar to the \hype benchmark, we reveal the correct answer to the participant upon submitting a response to every question. Every participant is allowed to participate in our study only once, which ensures unique responses across all experiment conditions. At the end of the study, we reveal the overall accuracy of the participant in our task and pay them \$2 for their time~\cite{hype}.\looseness=-1

\begin{table}[t]
	\small
	\centering
	\ra{1.5}
	\begin{tabularx}{0.95\columnwidth}{l X}
		\toprule
		Q1~~~~~ & What is your gender?\\
		Q2 & What is the highest level of education that you have completed?\\
		Q3 & What is your age?\\
		Q4 & Do you play video games?\\
		Q5 & How many hours per day do you play video games?
		\textit{\scriptsize (only asked if the participant plays video games)}\\
		\bottomrule
	\end{tabularx}
	\caption[User study questionnaire]{Pre-study questionnaire. Except for \textit{age}, all other questions are multiple-choice.}
	\label{chap-deepnag-tab-prequestions}
\end{table}

When posting our study on the Mechanical Turk platform, we created a list of criteria to ensure the selection of a pool of high-quality workers. We refined and validated these criteria through trials on Mechanical Turk prior to starting our actual study. First, participants must have an approval rating of at least 97\% to participate in our study to filter out low-quality workers. Our next participation requirement is that workers must have completed at least 5000 studies. This criterion filters out participants who may have high approval ratings because they recently joined the platform. Lastly, participants must be \textit{Mechanical Turk Masters} to be eligible to participate in our study. Amazon uses proprietary criteria to grant top-performing workers this qualification. Although the exact criteria is not publicly disclosed, Amazon claims they continuously monitor the performance of master workers across different user studies on the platform to ensure consistent performance\footnote{Details available at \url{https://www.mturk.com/worker/help}}.

Our study consisted in evaluating each of DeepGAN and DeepNAG on three datasets covering different gesture modalities: Kinect (JK-2017~\cite{jackknife}), Leap Motion (JK-2017~\cite{jackknife}) and Pen gestures (\$1-GDS~\cite{one_dollar}). Thus our \textit{generator} factor has two levels and our \textit{dataset} factor has three levels, yielding a total of six experiments.

\begin{figure*}[t]
	\centering
	\begin{tabular}{l l}
		\includegraphics[width=0.30\linewidth]{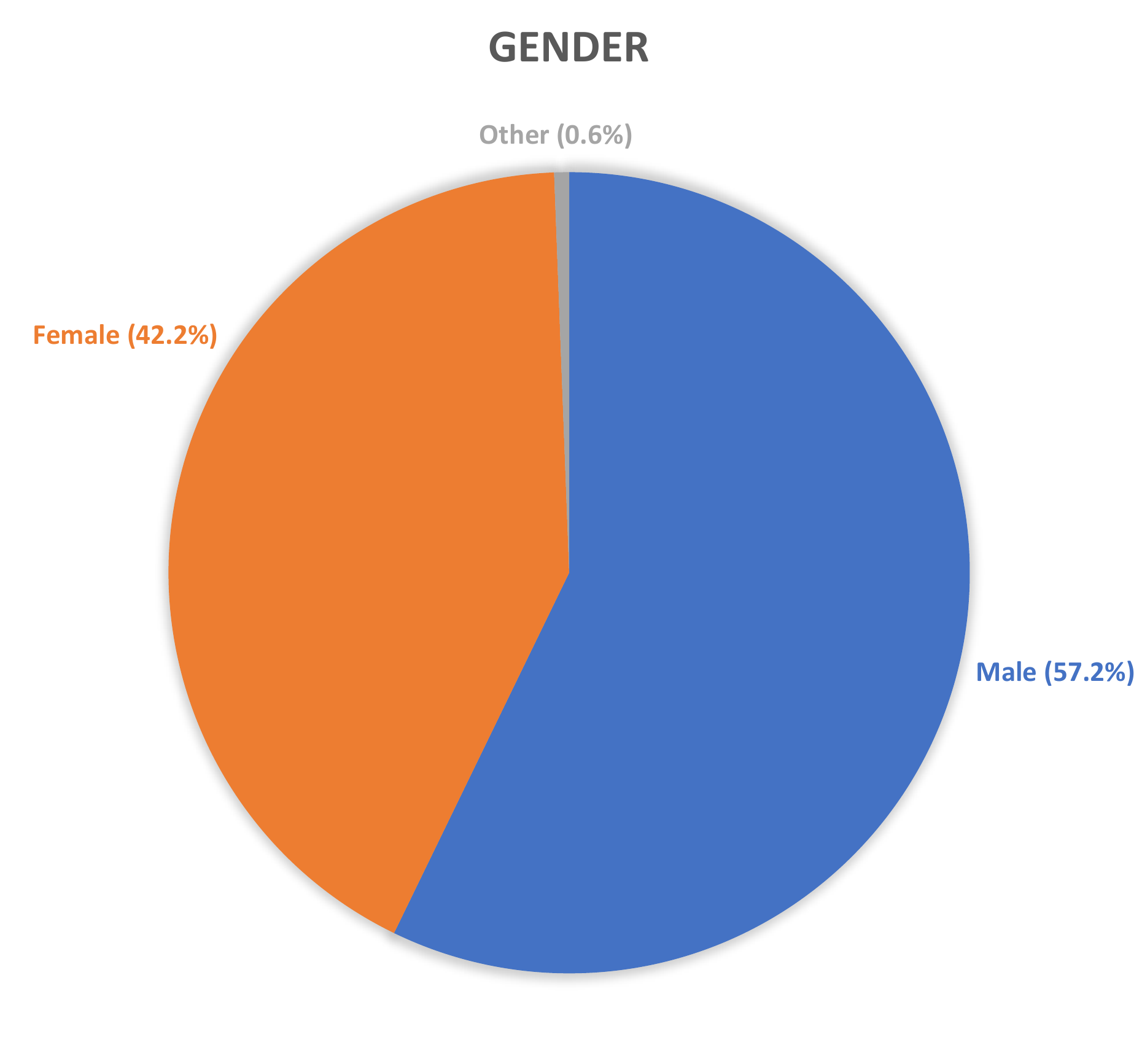} &
		\includegraphics[width=0.30\linewidth]{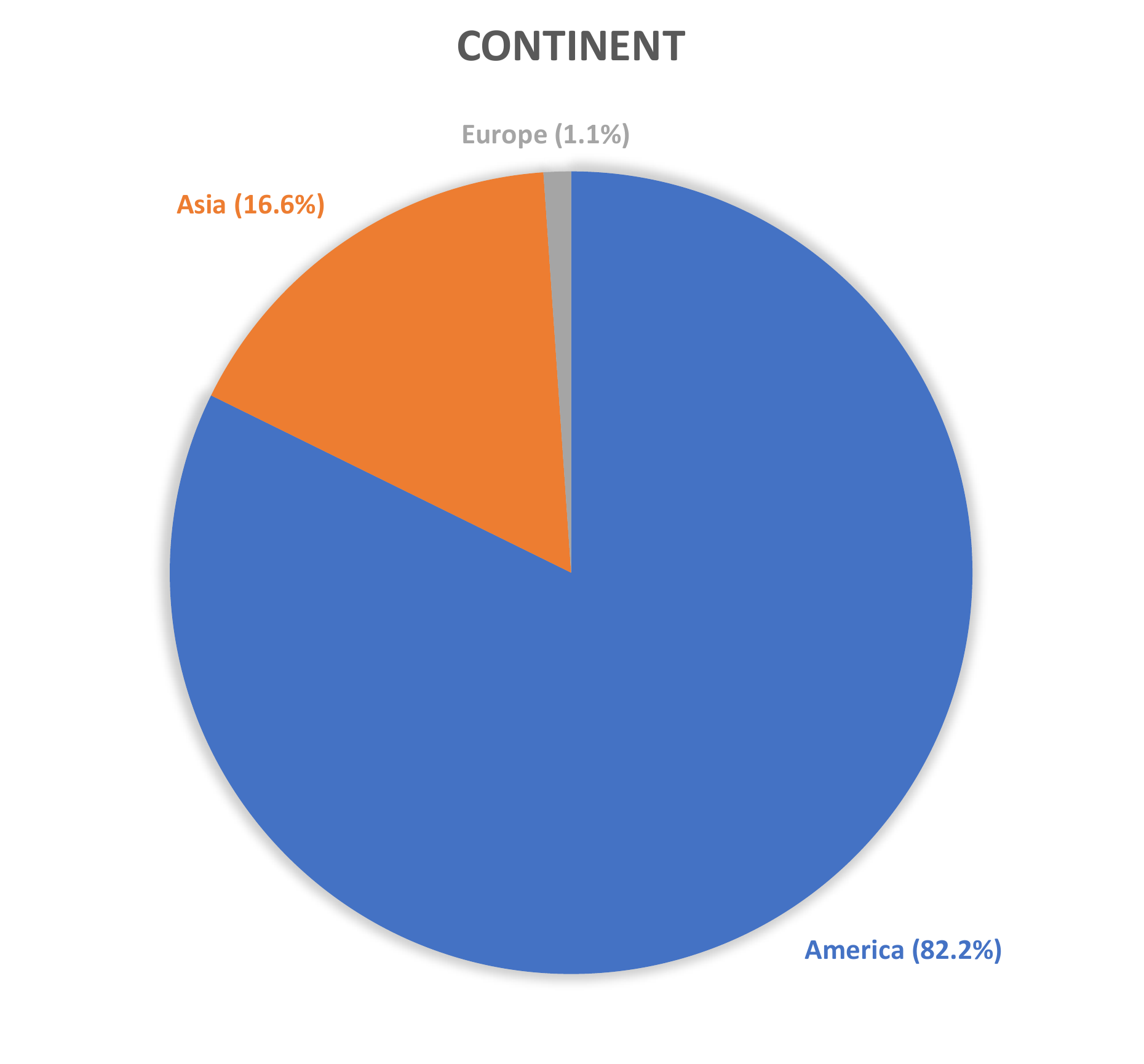} \\
		\includegraphics[width=0.30\linewidth]{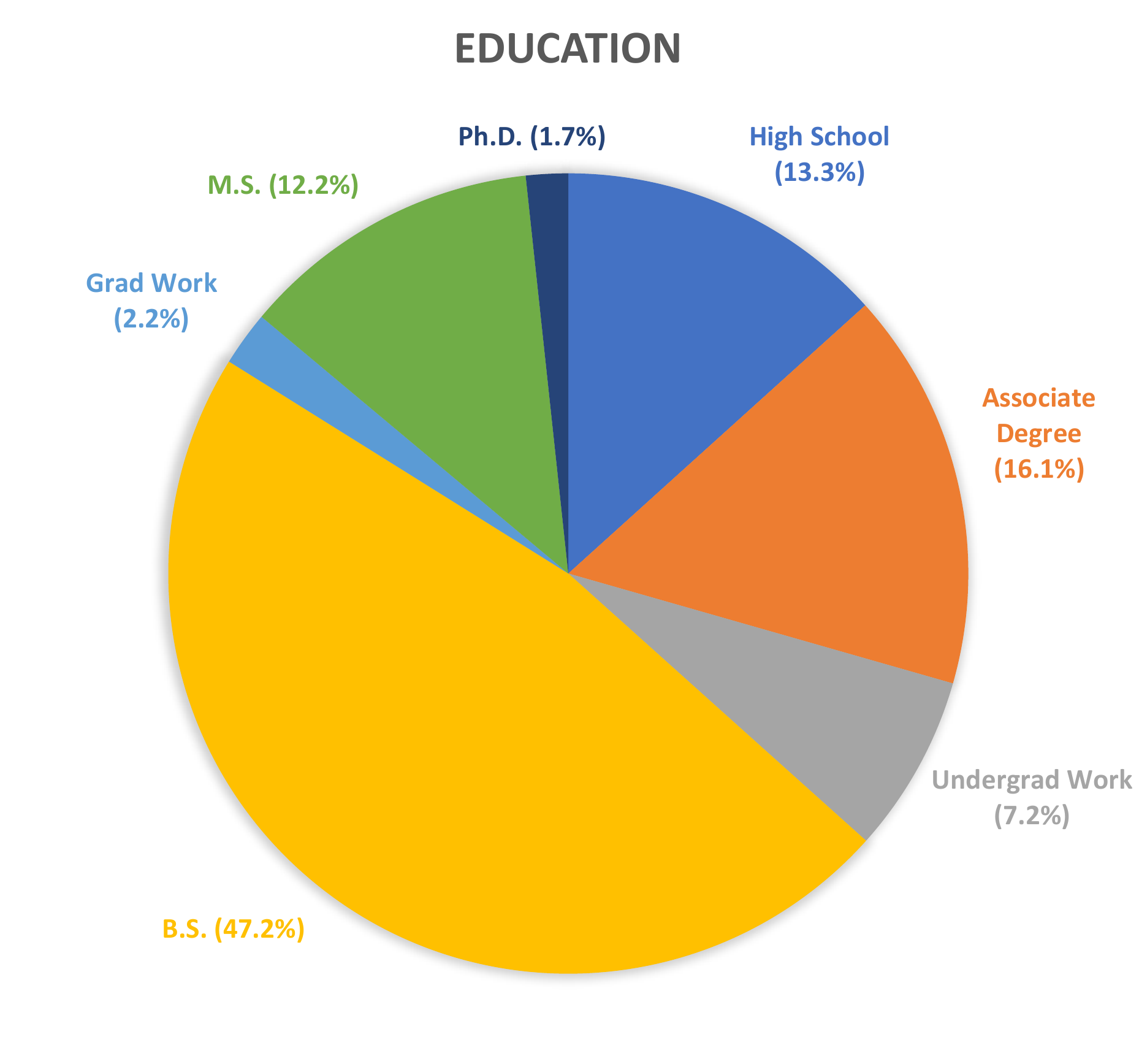} &
		\includegraphics[width=0.30\linewidth]{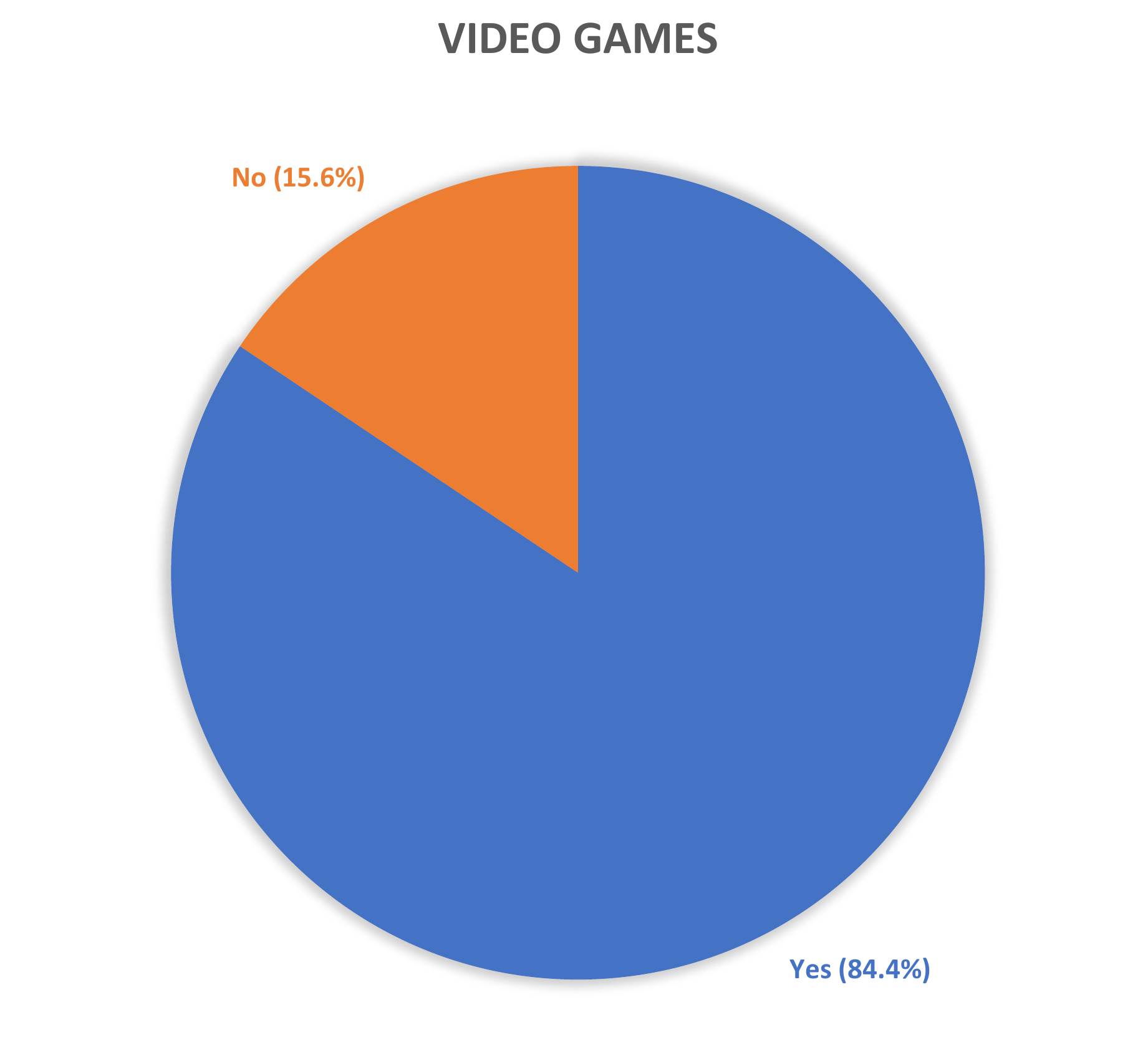}
	\end{tabular}
	\caption{Demographics of our user study participants (best viewed in color).}
	\label{chap-deepnag-fig-demographic}
\end{figure*}

\begin{table*}[tbh]
	\small
	\centering
	\ra{1.2}
	\begin{tabular}{l l d{3.2} d{3.2} d{3.2} d{3.2}}
		\toprule
		\textbf{Dataset} & \textbf{Generator} & \mc{\textbf{~~$\text{HYPE}_{\infty}$}} & \mc{\textbf{Std.}} & \mc{\textbf{Fake Errors}} & \mc{\textbf{Real Errors}}\\
		\midrule
		JK2017 (Kinect)~\cite{jackknife} & \textbf{DeepNAG} & 48.1 & 8.8 & 53.9 & 42.3\\
		& DeepGAN & 38.4 & 12.9 & 44.3 & 32.5\\
		\cmidrule(l{4pt}r{4pt}){2-6}
		JK2017 (LeapMotion)~\cite{jackknife}~~~ & \textbf{DeepNAG} & 51.0 & 4.3 & 56.1 & 45.8\\
		& DeepGAN & 22.7 & 11.4 & 23.9 & 21.4\\
		\cmidrule(l{4pt}r{4pt}){2-6}
		\$1-GDS~\cite{one_dollar} & \textbf{DeepNAG} & 50.0 & 6.7 & 56.3 & 43.7\\
		& DeepGAN & 44.4 & 8.3 & 49.5 & 39.3\\
		\bottomrule
	\end{tabular}
	\caption[User study results on Amazon Mechanical Turk]{Amazon Mechanical Turk user study results. Reported values are percentages (averaged over 30 participants). The top performing model on each dataset (with statistical significance) is boldfaced.}
	\label{chap-deepnag-tab-user-study}
\end{table*}

\vspace{1mm}
\noindent
\textbf{Results and discussion.~}
In total, we recruited 180 participants with an average age of 41 years ($\sigma$=10.8). Figure~\ref{chap-deepnag-fig-demographic} depicts the demographics of our participants. A majority of our participants indicated that that they played video games. Those who did, played an average of 2.2 hours per day ($\sigma$=1.8). Across all tasks, participants spent an average of 12.3 minutes ($\sigma$=3.8), and each question was answered in 7.4 seconds on average ($\sigma$=2.3). Considering a payment of \$2 per study, our participants were compensated well above the minimum wage specified by the United States federal guidelines (\$7.25 per hour at the time of this writing).

Table~\ref{chap-deepnag-tab-user-study} presents the results of our user study. In all experiments, we observe higher \hype scores for DeepNAG compared to DeepGAN. Unpaired t-tests confirm that the difference is significant in all experiments: $t$(58)$=$3.3, $p$=0.001 (JK2017-Kinect~\cite{jackknife}), $t$(58)=12.4, $p<$0.001 (JK2017-Leap Motion~\cite{jackknife}) and $t$(58)=2.8, $p$=0.006 (\$1-GDS~\cite{one_dollar}). 

Focusing on the results with JK2017 (Leap Motion)~\cite{jackknife} dataset, we observe a large \hype score gap between the two generators. Notably, DeepNAG achieves hyper-realism on this dataset: its fake samples look more realistic to humans than the real ones. These results correlate well with those in Section~\ref{sec-evaluation-aug}: on the Leap Motion dataset, DeepNAG significantly outperformed DeepGAN in reducing the recognition error (Table~\ref{tab-scores}) and in some cases, DeepNAG reduced the recognition error to zero (Figure~\ref{fig-deepgan-results}).

Similar to ~\cite{hype}, we report a breakdown of the error on the real and fake samples. We observe higher fake errors with DeepNAG in all cases. Inline with Zhou~\etal's observation~\cite{hype}, real and fake errors track each other. This indicates participants become more confused when fake samples are particularly hard to distinguish from the real ones.

To investigate whether there is an association between playing video games and the ability to distinguish between real and fake samples, we performed a multiple regression analysis using \textit{dataset}, \textit{generator} and \textit{play video games} as predictors. The results show that there is  no statistically significant association between playing video games and \textit{accuracy} when controlled for \textit{dataset} and \textit{generator} $\big($ coeff=0.01, $p$=0.51, CI (95\%)=(-0.029, 0.058) $\big)$.

In summary, our study shows that it is harder for evaluators to distinguish DeepNAG's synthetic samples from the real samples compared to those produced by DeepGAN. This trend holds regardless of the dataset and gesture modality. Additionally, DeepNAG not only outperformed DeepGAN in every experiment, but it also achieved hyper-realism on the Leap Motion dataset. These results correlate well with our data augmentation performance evaluations in Section~\ref{sec-evaluation-aug} and are notable considering that DeepNAG and DeepGAN both use the same underlying generator.

\section{Conclusion}
We discussed modality-agnostic gesture generation with recurrent neural networks. We first presented DeepGAN, our GAN model for synthetic gesture generation across various datasets and gesture modalities. To reduce the training complexity, we formulated a novel loss function based on the dynamic time warping (DTW) algorithm and the average Hausdorff distance. Our loss function obviated the need for a separate discriminator network, and led to 12--17$\times$ faster training. We called this approach DeepNAG and evaluated it from two aspects. Our first evaluations focused on the use of either model towards data augmentation for improved gesture recognition. In these evaluations, DeepNAG outperformed DeepGAN, along with other synthetic gesture generators across various datasets and recognizers. Next, we evaluated the perceived quality of the synthetic samples produced by DeepGAN and DeepNAG using human evaluators. Our user study, which was based on the \hype benchmark and was conducted using Amazon Mechanical Turk, demonstrated that DeepNAG consistently outperformed DeepGAN in terms of the realism of the synthetic samples. Users confused DeepNAG's samples with the real ones more frequently, and on one of our studied datasets, DeepNAG achieved hyper-realism by obtaining a \hype score of 51\%.\looseness=-1

In the future, we plan to more deeply explore the generation of multi-actor gestures, as well as the use of domain adaptation techniques to further improve data augmentation performance. Lastly, we aim to explore the application of our loss function in problem domains besides gestures such as time-series generation.

\appendix
\section{Appendix: Gesture Generation with Variational Autoencoders}
\label{appendix}
Thus far, we have shown that training our proposed RNN-based gesture generator using our novel loss function outperforms the same generator that is trained with the improved Wasserstein loss in a GAN training setting. In this section we investigate whether our generator can be trained in a variational autoencoder setting.\looseness=-1

\subsection{Background}
Variational autoencoders (VAE)~\cite{vae} are a class of autoencoders~\cite{autoencoder,autoencoder-hinton} designed for generative modeling. Similar to autoencoders, VAEs consist of encoder and decoder networks. However, the goal of VAEs is to model the distribution of the input data by learning a latent representation thereof. As such, the encoder network maps the input data $\xreal$ to a probability distribution (latent space) while the decoder network aims to reconstruct the original data from a vector $\latent$ in that latent space. Once training concludes, the decoder network can be used to generate synthetic samples. The loss function for VAEs consists of reconstruction as well as regularization terms. The reconstruction term ensures that the reconstructed data closely resembles the input data. The regularization term ensures that the learned distribution of the latent space is as close to some known distribution as possible (typically the standard normal distribution). Assuming that $\phi$ and $\theta$ denote the trainable parameters of the encoder and decoder respectively, the following is the loss function that is minimized~\cite{vae-loss}:

\begin{align}
\label{chap-background-eq-vae}
\loss(\theta, \phi; \xreal, \latent)~=~&
\underbrace{\mystrut{3ex}
	-\mathbb{E}_{q_{\phi}(\latent \lvert \xreal)} \big\lbrack \text{log}~p_{\theta}(\xreal | \latent) \big\rbrack
}_{\text{reconstruction}}
~+~\\
&
\underbrace{\mystrut{3ex}
	D_{KL} \big( q_{\phi}(\latent \lvert \xreal) ~\big\lvert\big\rvert~ p(\latent) \big)
}_{\text{regularization}} \nonumber
\end{align}

\noindent
where $D_{KL} (~\big\lvert\big\rvert~)$ is the Kullback-Leibler (KL) divergence between two probability distributions. Equation~\ref{chap-background-eq-vae} simply aims to minimize the reconstruction error as well as the KL divergence between the learned latent space and the standard normal distribution ($p(\latent) = \mathcal{N}(0, 1)$). Note that in this formulation, class labels are not considered, which means one cannot control what sample class is produced for a given $\latent$. Sohn~\etal~\cite{cvae} proposed conditional VAEs in which a conditioning criteria is applied to the input data $\xreal$ as well as the latent vector $\latent$ similar to conditional GANs as described in Section~\ref{sec-gan}.

\subsection{Model Architecture and Objective Function}
We iteratively designed our VAE's overall architecture. Given our goal of training the generator of DeepGAN/DeepNAG in a VAE framework, we reused the aforesaid generator as the decoder in our VAE network. As for the encoder, we started with using the uDeepGRU model as the encoder. This way, our overall VAE network closely resembled that of DeepGAN's. After running some preliminary experiments, we observed that the choice of the encoder architecture did not result in perceptible difference in the model's performance. In fact, adding or removing layers in either the encoder or the decoder made little difference in the produced results, inline with what Bowman~\etal~\cite{sentence-vae} observed. We ultimately decided to carry on with an architecture similar to Figure~\ref{fig-gan}.

We now discuss our proposed training objective function which can be used to train our RNN-based generator in a VAE framework. As previously mentioned, the VAE objective function typically consist of reconstruction and regularization terms. We can conveniently reuse the regularization term of Equation~\ref{chap-background-eq-vae}, as it simply ensures that the learned latent space follows the standard normal distribution. The reconstruction term, however, is domain-specific. To our knowledge, no reconstruction loss term for generating gestures has been previously discussed in the literature. Recall that the this term ensures that the output of the decoder (generator) closely resembles the input data. Conveniently, a differentiable metric that can be used for this purpose is sDTW. Thus, we propose the following loss function as the objective to minimize during the training our gesture generating VAE:

\begin{align}
\label{chap-deepnag-eq-vaeg}
\loss(\theta, \phi; \xreal, \latentcond)~=~&
\text{sDTW}\big( \xreal, G_{\theta}(\latentcond); f \big)
~+~\\
& D_{KL} \big( q_{\phi}(\latentcond \lvert \xreal) ~\big\lvert\big\rvert~ p(\latentcond) \big) \nonumber
\end{align}

\noindent
where $f$ is sDTW's internal cost function and $\latentcond$ is the latent vector conditioned on the class label $c$. In simple terms, we define the reconstruction error as the sDTW dissimilarity between the input data and the output of the generator. As mentioned in Section~\ref{sec-nag}, Cuturi and Blondel show that the computed sDTW value would be the expected value of dissimilarity between two time-series, over every possible alignment between them weighted by their probability under the Gibbs distribution~\cite{sdtw}. This closely resembles the original reconstruction term of VAEs (Equation~\ref{chap-background-eq-vae}). To our knowledge, our proposed loss function is novel, and we are the first to formulate such function for gesture generation using VAEs.

\subsection{Results}
As previously mentioned, changes in the architecture of our model made little difference in the produced results. Although training plots showed a steady decrease of the loss value and training converged, the generated results lacked visual quality and diversity. Most gesture trajectories were rather noisy. Again, we observed this trend regardless of the architecture of encoder or decoder networks, various hyperparameter settings, the choice of $f$ (we tried ED and COS) or even the gesture dataset. We additionally experimented with alternative reconstruction terms. Specifically, we experimented with the mean squared error (MSE) of both the Euclidean distance as well as the cosine similarity of gesture paths ($\xrealvec$) between the input and reconstructed samples. These alternate formulations performed worse than our sDTW-based reconstruction term.\looseness=-1

\begin{figure}[t]
	\centering
	\begin{tabular}{c}
		\includegraphics[width=0.99\linewidth]{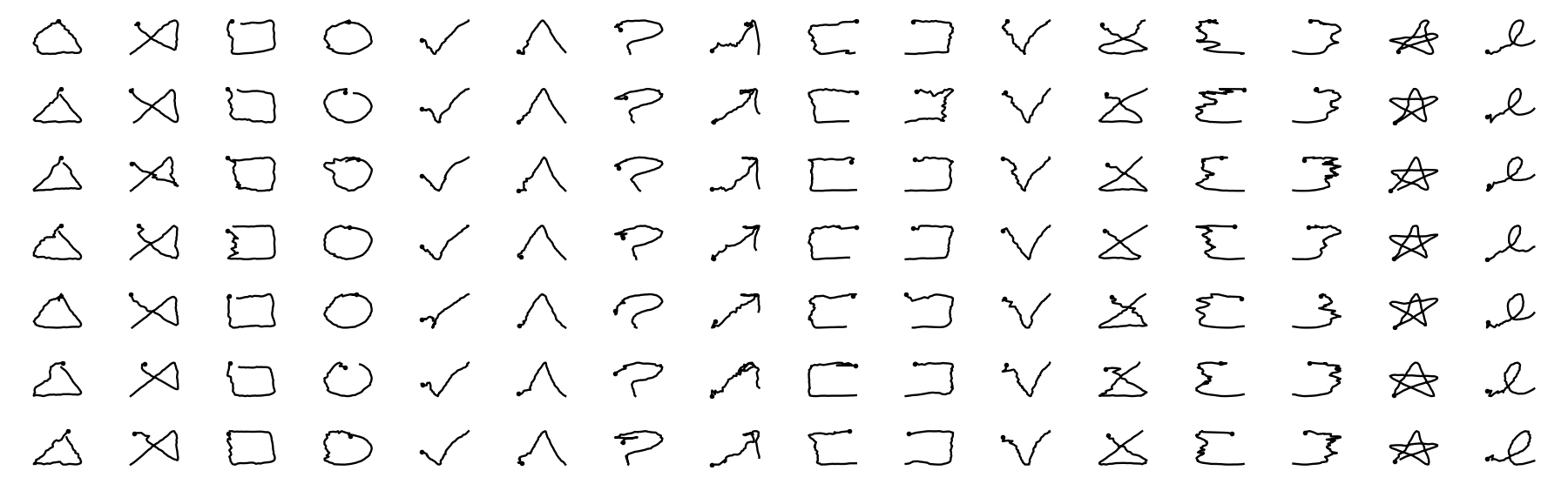}\\
		{\footnotesize (a) VAE (samples)}\\
		~\\
		\includegraphics[width=0.99\linewidth]{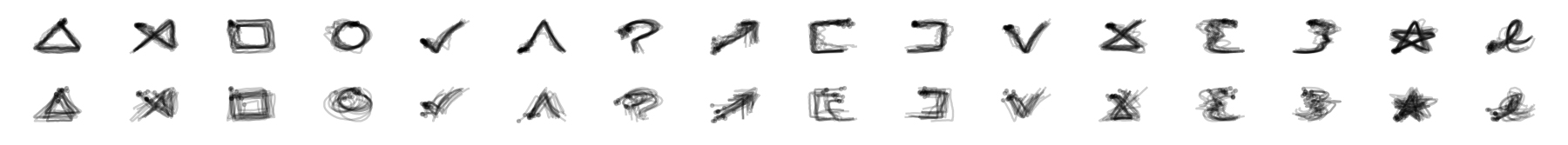}\\
		{\footnotesize (b) VAE (overlays)}
	\end{tabular}
	\caption[VAE generation results]{Synthetic gestures produced by our VAE-based generator trained on \$1-GDS~\cite{one_dollar} dataset. Note that most samples are noisy and lack visual fidelity. We further show overlayed rendering of synthetic samples from our VAE model (b -- top), and real samples (b -- bottom). Each overlay consists of 16 samples per class. Note the lack of variety in the synthetic results compared to the real samples.}
	\label{chap-deepnag-fig-vae-results}
\end{figure}

Some samples produced by our VAE model when trained on the \$1-GDS dataset~\cite{one_dollar}, along with overlayed samples for each of the real and synthetic data are depicted in Figure~\ref{chap-deepnag-fig-vae-results}. These results show that the produced samples lack sufficient diversity when compared to the real samples. Given the visual quality of the results, we hypothesize that our proposed VAE framework is not suitable for training our generator to produce good synthetic samples.

{\small
\bibliographystyle{ieee_fullname}
\bibliography{bibliography}
}

\end{document}